\documentclass[10pt,twocolumn,letterpaper]{article}

\usepackage[pagenumbers]{cvpr}

\usepackage[table]{xcolor}
\usepackage{multirow}
\usepackage{graphicx}
\usepackage{amsmath}
\usepackage{amssymb}
\usepackage{booktabs}

\usepackage[pagebackref,breaklinks,colorlinks]{hyperref}

\usepackage[capitalize]{cleveref}
\crefname{section}{Sec.}{Secs.}
\Crefname{section}{Section}{Sections}
\Crefname{table}{Table}{Tables}
\crefname{table}{Tab.}{Tabs.}

\def\cvprPaperID{39}
\def\confName{CVPR}
\def\confYear{2023}

\begin{document}

\title{NutritionVerse-3D: A 3D Food Model Dataset for Nutritional Intake Estimation}

\author{
Chi-en Amy Tai\textsuperscript{1}
\qquad Matthew Keller\textsuperscript{1}
\qquad Mattie Kerrigan\textsuperscript{1}
\qquad Yuhao Chen\textsuperscript{1}
\qquad Saeejith Nair\textsuperscript{1} \\
\qquad Pengcheng Xi\textsuperscript{2}
\qquad Alexander Wong\textsuperscript{1} \\
\textsuperscript{1} Vision and Image Processing Lab, University of Waterloo\\
\textsuperscript{2} National Research Council Canada \\
{\tt\small \{amy.tai, m6keller, makerrig, yuhao.chen1, smnair, alexander.wong\}@uwaterloo.ca} \\
{\tt\small {pengcheng.xi}@nrc-cnrc.gc.ca}
}

\maketitle

\begin{abstract}
77\% of adults over 50 want to age in place today, presenting a major challenge to ensuring adequate nutritional intake. It has been reported that one in four older adults that are 65 years or older are malnourished and given the direct link between malnutrition and decreased quality of life, there have been numerous studies conducted on how to efficiently track nutritional intake of food. Recent advancements in machine learning and computer vision show promise of automated nutrition tracking methods of food, but require a large high-quality dataset in order to accurately identify the nutrients from the food on the plate. Unlike existing datasets, a collection of 3D models with nutritional information allow for view synthesis to create an infinite number of 2D images for any given viewpoint/camera angle along with the associated nutritional information. In this paper, we develop a methodology for collecting high-quality 3D models for food items with a particular focus on speed and consistency, and introduce NutritionVerse-3D, a large-scale high-quality high-resolution dataset of 105 3D food models, in conjunction with their associated weight, food name, and nutritional value. These models allow for large quantity food intake scenes, diverse and customizable scene layout, and an infinite number of camera settings and lighting conditions. NutritionVerse-3D is publicly available as a part of an open initiative to accelerate machine learning for nutrition sensing. 
\end{abstract}

\section{Introduction}
\begin{figure}[h]
    \begin{center}
        \includegraphics[width=\linewidth]{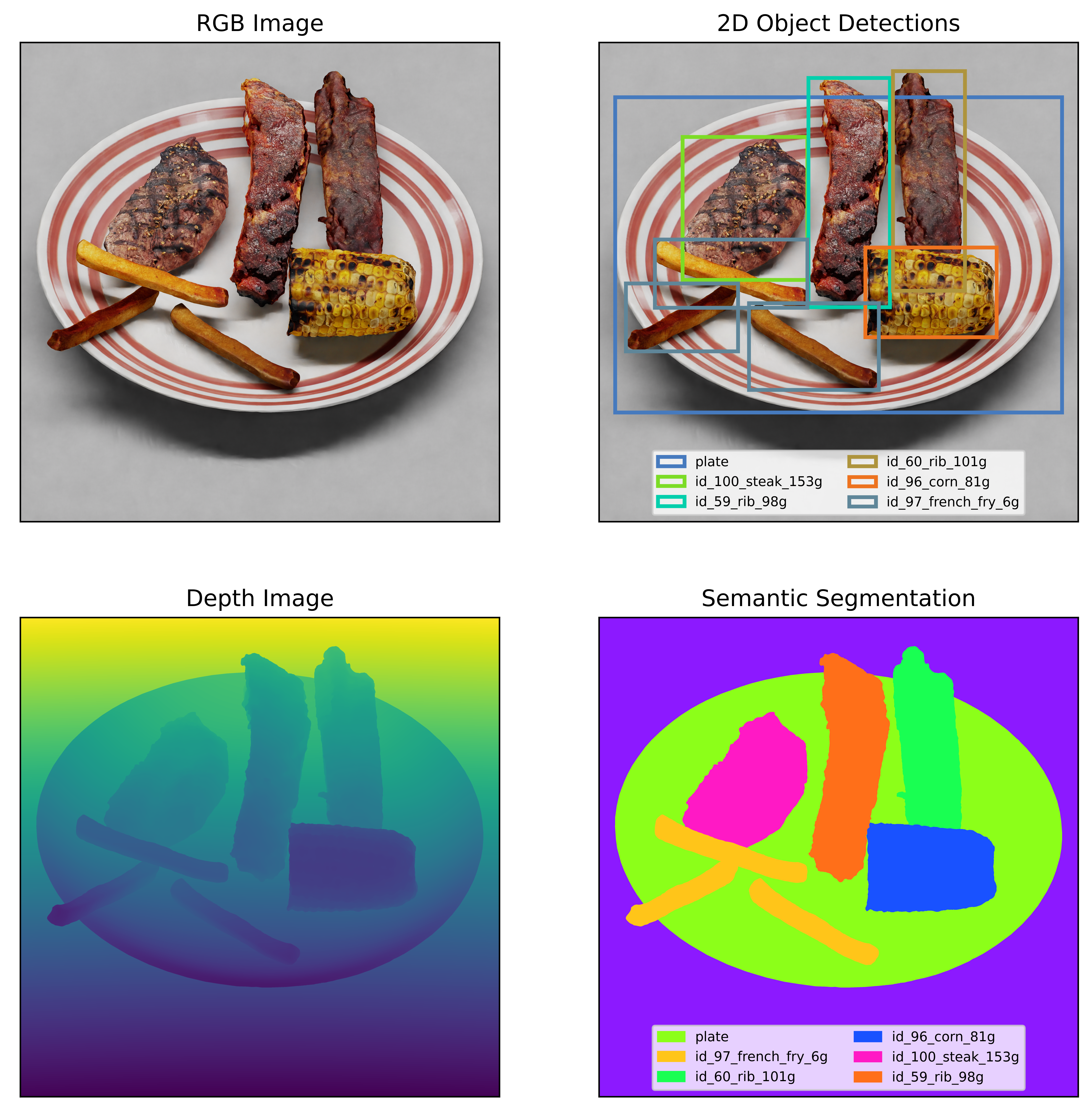}
    \end{center}
    \caption{Example of a 3D food scene with multi-modal image data and annotation metadata that can be automatically generated using the proposed NutritionVerse-3D dataset. top-left: RGB image, bottom-left: depth image, top-right: object detection annotations, bottom-right: semantic segmentation annotations.}
    \label{fig:curated-3d-scene}
\end{figure}
\begin{figure}[h]
    \begin{center}
        \includegraphics[width=\linewidth]{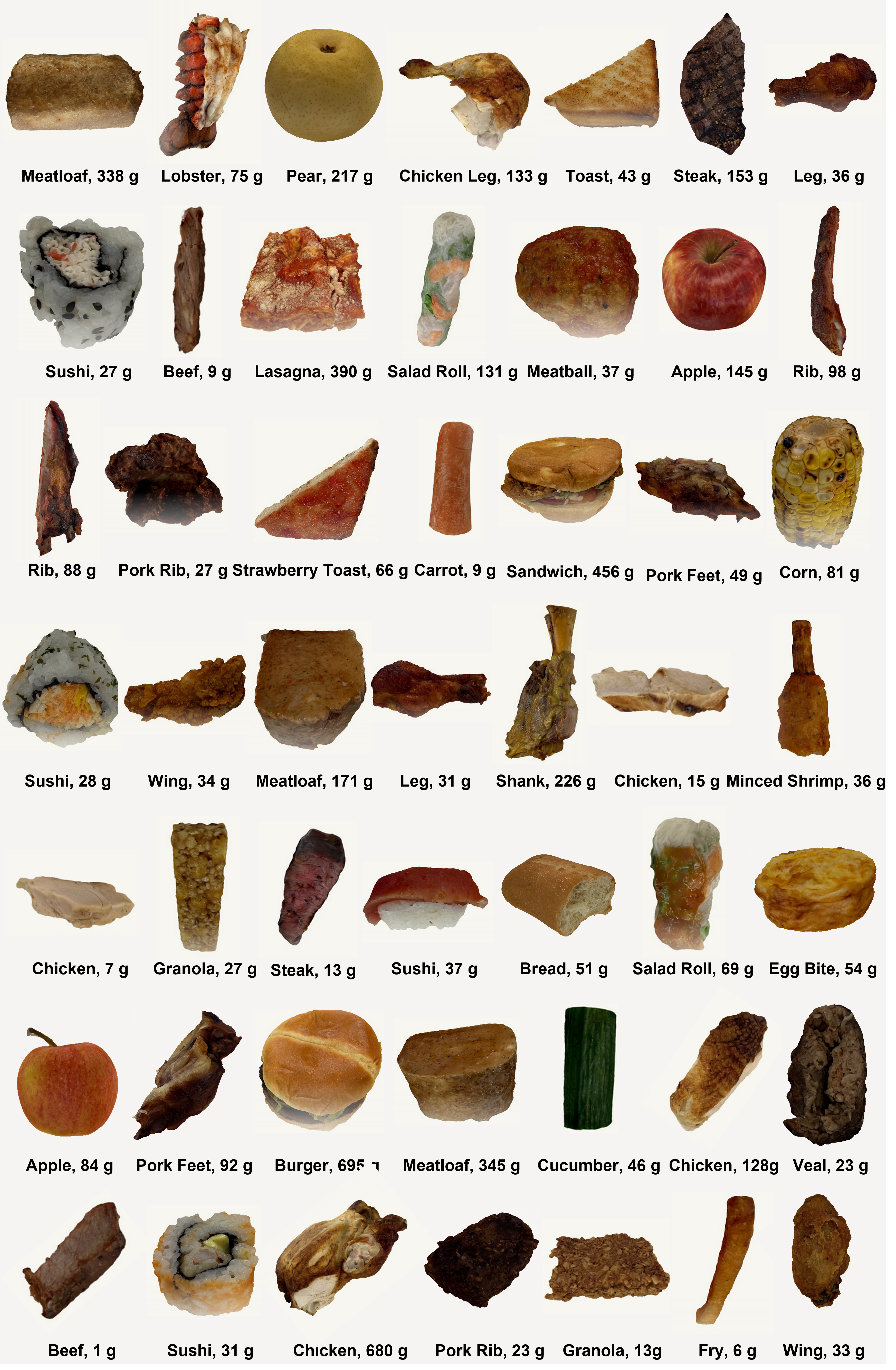}
    \end{center}
    \caption{Examples of 3D models in the proposed NutritionVerse-3D dataset.}
    \label{fig:model-grid}
\end{figure}

\begin{figure*}
    \begin{center}
        \includegraphics[width=\linewidth]{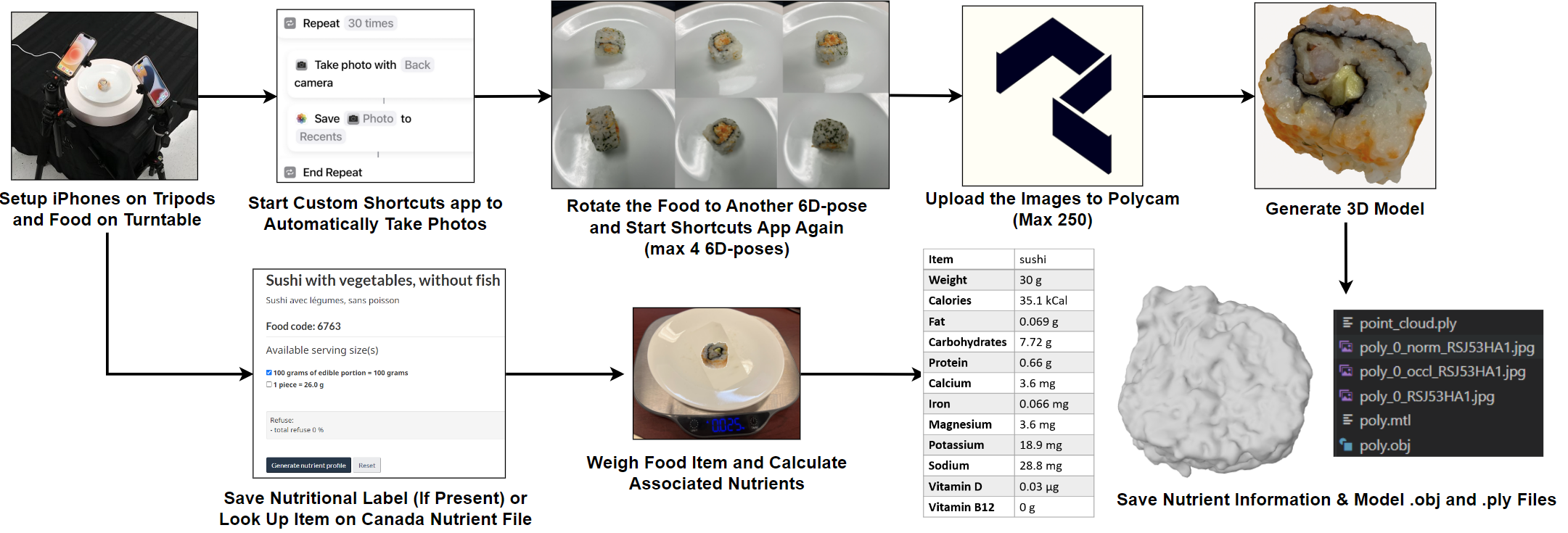}
    \end{center}
    \caption{Overview of the entire data collection and creation process for NutritionVerse-3D.}
    \label{fig:entire-process-map}
\end{figure*}

\begin{figure}
    \begin{center}
        \includegraphics[width=\linewidth]{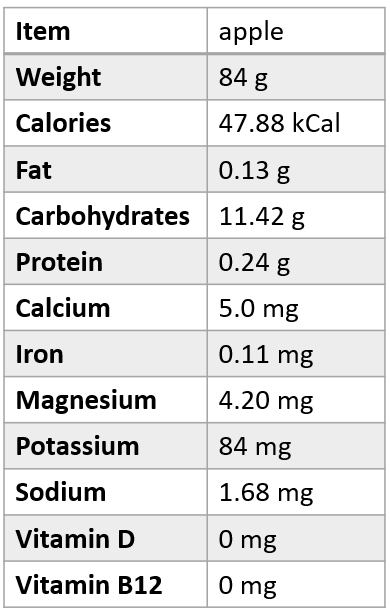}
    \end{center}
    \caption{Example of the nutritional information given for each 3D model in the NutritionVerse-3D dataset (apple).}
    \label{fig:nutrition-ex}
\end{figure}

The desire to age in place has grown immensely in the past decade with 77\% of adults over 50 wanting to stay at home in 2021 ~\cite{AIP-Stats}. However, one of the main challenges with aging in place is ensuring adequate food nutritional intake. It has been reported that one in four older adults that are 65 years or older are malnourished ~\cite{malnutrition-frequency}. Given the direct link between malnutrition and decreased quality of life ~\cite{malnutrition-qol}, there have been numerous studies conducted on how to efficiently track food nutritional intake. Recent advancements in machine learning and computer vision show promise of automated nutrition tracking methods of food \cite{food-recog-promise, imaging-volm-promise}, but require a large high-quality dataset in order to accurately identify the nutrients from the food on the plate. Unfortunately, existing datasets such as FoodSeg \cite{wu2021foodseg} and Nutrition5k \cite{thames2021nutrition5k} do not have per-item nutrition details. Existing datasets also comprise of 2D images with fixed or randomly selected camera views that are discretely sampled \cite{food-recog-promise, foodx-251, uecfood-100-dataset, food2k-dataset, chinesefoodnet, food-101}. These set views introduce bias and are not very representative of the types of angles and quality of photos taken by older adults in realistic scenarios which would affect the training and accuracy of the model. 

Though the manual creation of a large-scale dataset that captures all potential angles and photo quality would be too time-consuming, leveraging view synthesis for 3D models with nutritional information can generate an infinite number of 2D images taken from any angle along with the associated nutritional information to reduce imbalance or bias towards a certain viewing angle. As such, by using view synthesis for 3D models, an infinite number of 2D images can be generated for any given viewpoint/camera angle. As shown in \cref{fig:curated-3d-scene}, using photorealistic 3D models allows for the creation of multi-modal datasets (RGB, depth) with photorealistic 2D images of meals from infinite possible food combinations, placements, and camera angles along with full annotiations (object annotations, segmentation annotations, etc.). 3D models also allow for large quantity food intake scenes, diverse and customizable scene layout, and an infinite number of camera settings and lighting conditions. In this paper, we develop a methodology for collecting quality 3D models for food items with a particular focus on speed and consistency, and introduce NutritionVerse-3D, a large-scale high-quality high-resolution dataset of 105 3D food models, in conjunction with their associated weight, food name, and nutritional value. Examples of the collected 3D models can be seen in \cref{fig:model-grid} with the methodology shown in \cref{fig:entire-process-map} and an example of the nutritional information available for each model shown in \cref{fig:nutrition-ex}. The NutritionVerse-3D dataset has been made publicly available \footnote{https://www.kaggle.com/datasets/amytai/nutritionverse-3d} as a part of an open initiative to accelerate machine learning for nutrition sensing. 

\section{Methodology}

The two primary factors considered in the design of the data collection pipeline are speed and consistency. Speed is important to maximize the number of food models that can be collected in a feasible amount of time for a large-scale dataset. Likewise, consistency is also critical to minimize human interaction and likelihood of variation in collecting data so that the number of high-quality food models obtained is optimized. 

Though it is now feasible to use automated wearable cameras, these devices have been found to be incredibly intrusive \cite{KELLY2013314} and pose significant ethical ramifications \cite{wearable-camera-ethics}. Given that the main goal is convenient nutritional intake tracking for older individuals, recent advances in mobile phone applications \cite{mobile-phone-applications, snap-and-eat} demonstrate that nutritional intake tracking through mobile devices would be more convenient and accepted by older individuals. Subsequently, mobile devices were chosen for collecting images and specifically, the iPhone \cite{apple-iphone} was chosen as the primary image capturing device due to its popularity and quality camera resolution (though any phone with a suitable camera could be used too). 

\begin{figure}[h]
    \begin{center}
        \includegraphics[width=\linewidth]{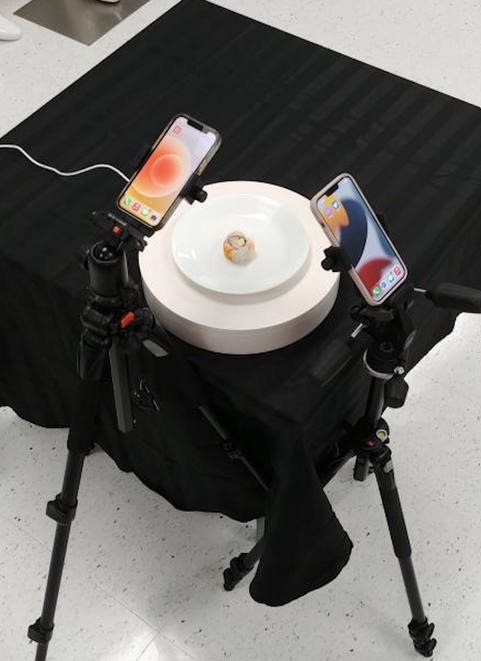}
    \end{center}
    \caption{Setup for the data collection process for an exemplar sushi piece.}
    \label{fig1}
\end{figure}

To generate a quality 3D model of a food item, various 3D scanner applications were compared based on their review rating, exporting capabilities, and ease of usage. In addition to having a high review rating and a variety of model export formats, Polycam \cite{polycam} also has a web interface with a shareable account for easy image input captured from multiple devices \cite{best_3d_scanner_apps}. Hence, leveraging the Polycam app, 3D models of food items are generated from 2D images taken by the iPhone. Consequently, three main restrictions are imposed by using the Polycam app. First, at least 70\% overlap between the photos is needed to produce quality 3D food models without holes or blurs. Second, a variety of angles of the food need to be captured to render a full model, and third, a maximum limit of 250 images is allowed for each food item.

To address the first restriction, an electric turntable with the default rotation speed of a full rotation in 24 seconds and a custom image taking script implemented using the built-in Shortcuts iPhone application is used to automatically collect consistent images of each food item in a short period of time whilst allowing for at least 70\% overlap between the photos. 

However, to meet the second limitation, a variety of angles need to be obtained for each food item. To ensure consistency between item captures, the camera angles and food 6D-poses collected for each food item should be the same. In experimenting with the number of camera angles, faster and more consistent performance is obtained using two iPhones set at two different angles compared to only one iPhone. Unfortunately, using two iPhones causes shadow interference in the image captures for each iPhone due to the lighting conditions in the room. In particular, the room has sparse fluorescent ceiling lights that are about 1 meter apart from each other. Therefore, we experimented with a variety of tripod layouts to discern the setup with the least amount of shadows on the turntable. As seen in Fig.~\ref{fig1}, the setup for the data collection process has two iPhones on adjacent tripods with very specific tripod distances for each iPhone and low shadow interference on the exemplar sushi piece on the turntable. 

\begin{figure}[h]
    \begin{center}
        \includegraphics[width=\linewidth]{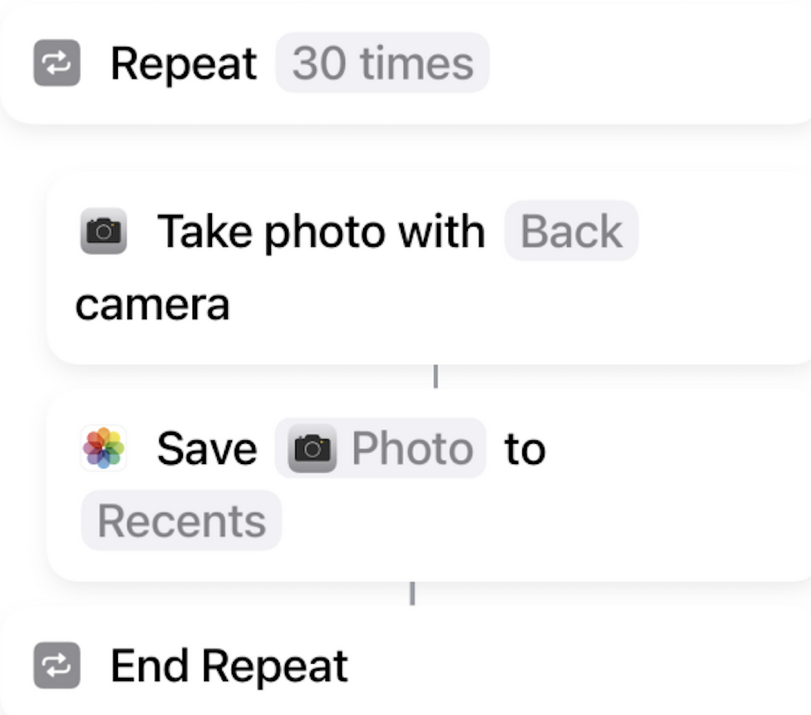}
    \end{center}
    \caption{Custom Shortcuts app used to take photos on the iPhones for data collection.}
    \label{fig2}
\end{figure}

In terms of the third main Polycam limitation, coordination between the number of photos taken and the combinations of the food item 6D-pose and the camera angle had to be determined. With a limit of 250 photos, the ideal scenario for data collection is to position the food in four different ways with two different camera angles. As such, the photo limit and the number of combinations leads to roughly 30 photos per food 6D pose-camera angle combination for a total of 240 images. Hence, as seen in the custom Shortcuts app in Fig. ~\ref{fig2}, the iPhones are configured to automatically take 30 consecutive photos. After taking 60 photos of the food item on one 6D-pose (30 photos per iPhone-tripod), the food item is rotated to another 6D-pose and the custom Shortcuts app is started again. Occasionally, due to the shape of some food items, four different food 6D-poses is  infeasible. For example, the egg and cheese bite could not stand on its side without rolling when the turntable rotated. Thus, to ensure consistency between image captures, the number of camera angles is increased to compensate for the lower number of possible food 6D-poses as seen in Table ~\ref{image-capture-settings}. 

\begin{table}[th]
    \caption{Overview of the food 6D-poses and camera settings combinations in data collection to produce a total of 240 images for the first and last row and a total of 180 images for the second row.}
    \setlength{\tabcolsep}{4pt}
    \centering
    \scalebox{1}{
        \begin{tabular}{|c|c|}
            \hline 
             \textbf{Num of Food 6D-poses} & \textbf{Num of Camera Angles}
            \\ 
            \hline
            2 & 4 \\
            3 & 2 \\
            4 & 2 \\
        \hline
        \end{tabular}}
    \label{image-capture-settings}
\end{table}

Though the setup led to successful 3D model renderings, these models often had pieces of their background included in the model itself. To address this problem, the object masking feature in the Polycam app is used to remove the background from the images and render only the food item. After conducting several experiments using plates with different textures or colours, it was determined that placing the food item on a white plate with low reflectivity and having a black tablecloth on top of the table rendered the most consistent quality of 3D models. Though the turntable has a white colour, the food item is not placed directly on the electric turntable as cleaning the turntable is risky and hence, may result in irreparable damage. 

\begin{figure}[h]
    \begin{center}
        \includegraphics[width=\linewidth]{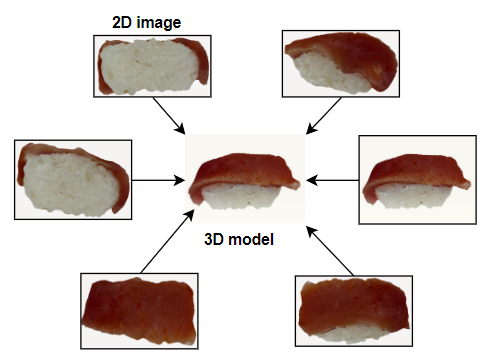}
    \end{center}
    \caption{Overall process to generate 3D models of food items in NutritionVerse-3D.}
    \label{process-chart}
\end{figure}

\begin{figure}[!ht]
    \begin{center}
    \subfloat[\centering 3D Model Image]{{\includegraphics[width=.4\linewidth]{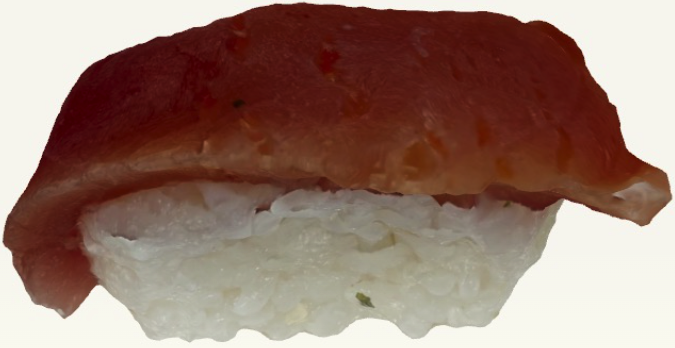} }}%
    \qquad
    \subfloat[\centering 3D Model Mesh]{{\includegraphics[width=.4\linewidth]{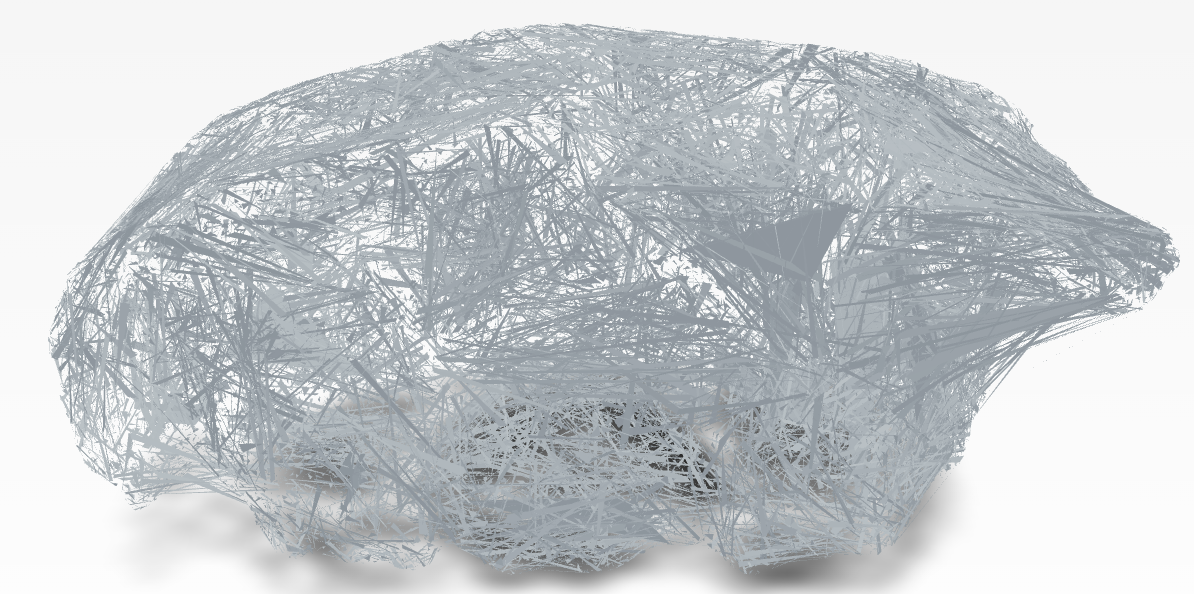} }}%
    \end{center}
    \caption{Example of a successful 3D model Polycam rendering.}
    \label{successful-rendering}
\end{figure}

\begin{figure}[h]
    \begin{center}
        \includegraphics[width=\linewidth]{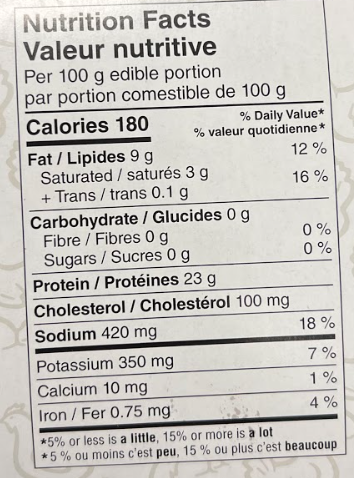}
    \end{center}
    \caption{Sample nutritional label for a food item used to record nutritional information.}
    \label{nutritional-label-example}
\end{figure}

\begin{figure}[h]
    \begin{center}
        \includegraphics[width=\linewidth]{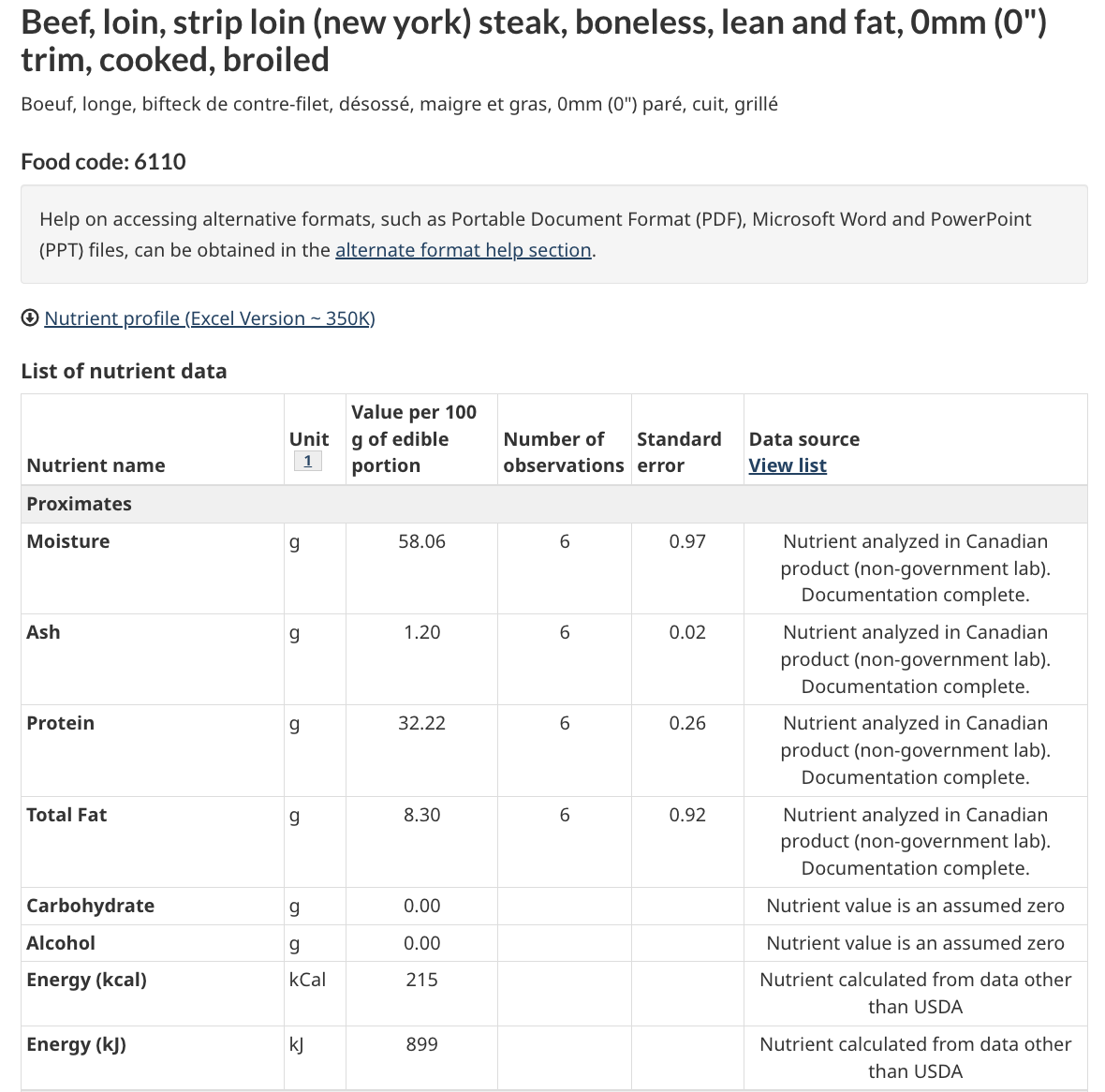}
    \end{center}
    \caption{Sample nutritional information from the Canada Nutrient File for a food item \cite{canada-nutrient-file}.}
    \label{govt-canada-example}
\end{figure}

The overall process to generate the 3D models of food items is shown in Fig.~\ref{process-chart} with an example of a successful 3D model rendering displayed in Fig.~\ref{successful-rendering}. The total weight and protein weight of each food item is weighed using a food scale and the food name is recorded for each food item. The nutritional value is obtained from the food packaging or from the Canada Nutrient File posted on the Government of Canada website \cite{canada-nutrient-file} for non-packaged food items such as apples. An example of a collected nutritional label on the food packaging is shown in Fig.~\ref{nutritional-label-example} and an example of the data available on the Government of Canada website for an example food item is shown in Fig.~\ref{govt-canada-example}.  

\subsection{Item-Specific Challenges}
\begin{figure}
    \begin{center}
        \includegraphics[width=\linewidth]{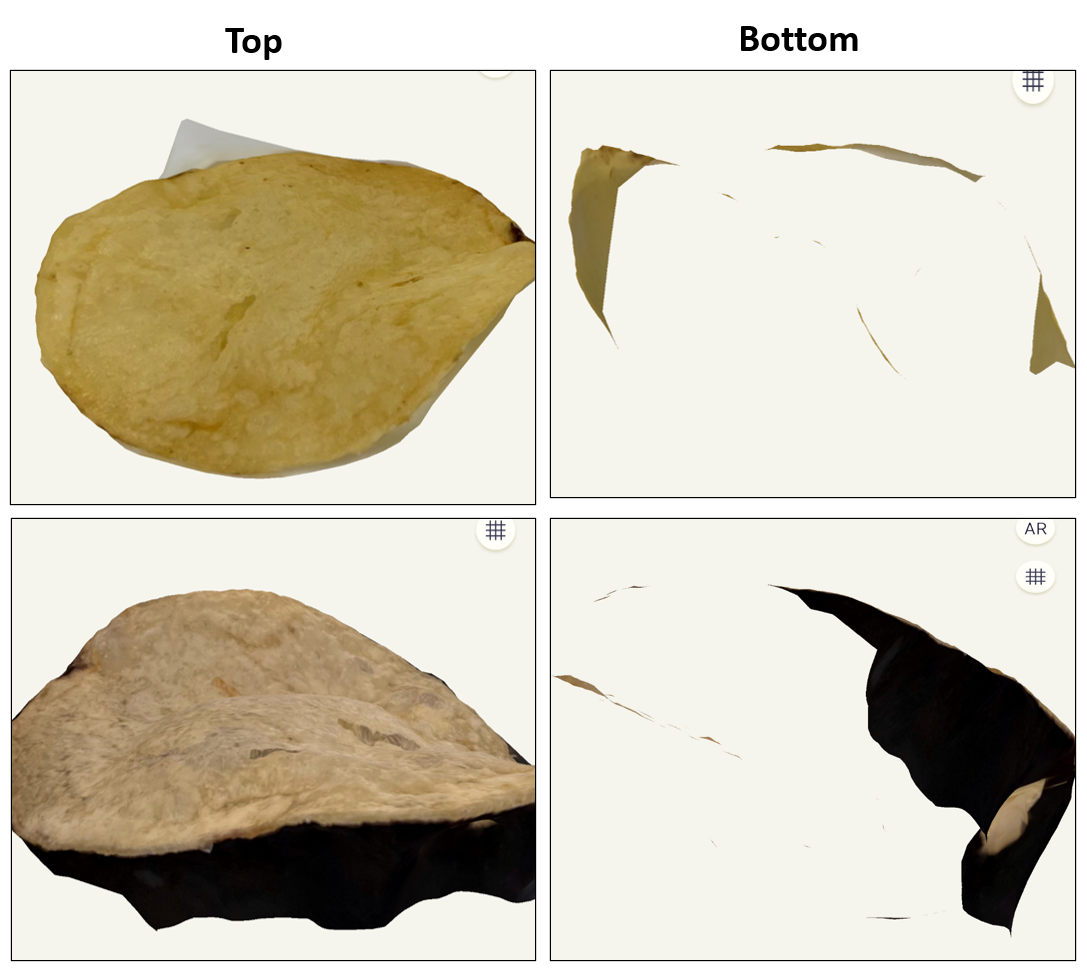}
    \end{center}
    \caption{Problematic renders of chips with holes on the bottom.}
    \label{fig:chip-issue}
\end{figure}

\begin{figure}
    \begin{center}
        \includegraphics[width=\linewidth]{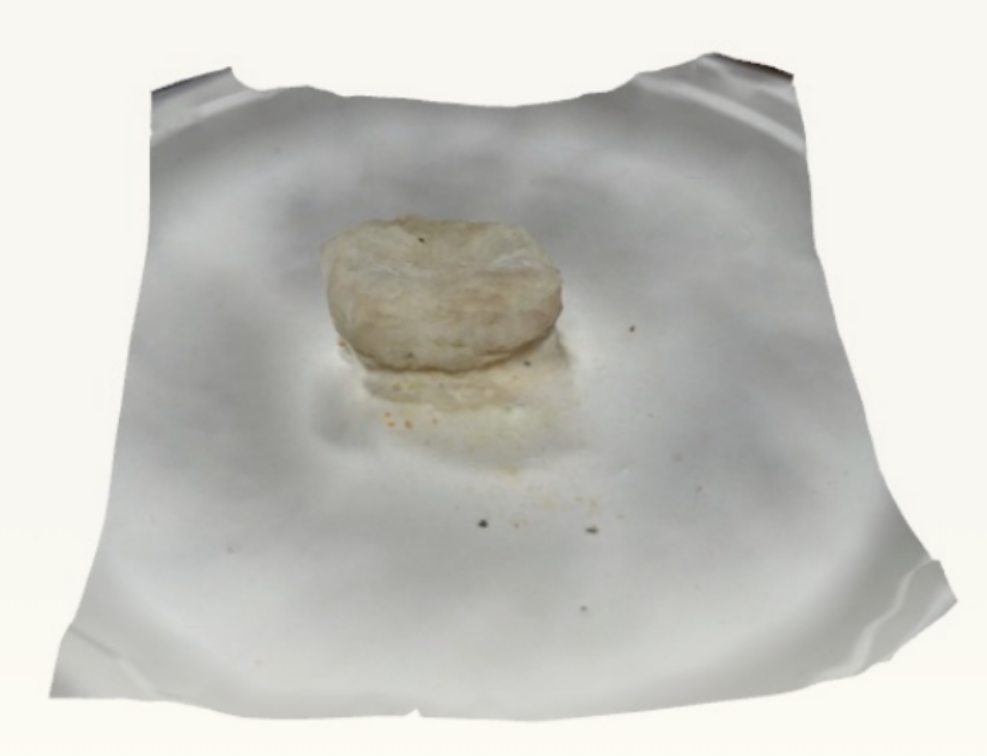}
    \end{center}
    \caption{Problematic renders of tuna rice ball due to high fragility.}
    \label{fig:tuna-issue}
\end{figure}

In the collection of various food items, we quickly discovered that it is easier to render 3D models of certain types of food compared to others. Specifically, models for textureless food such as cheese, thin foods such as chips, and small items such as grapes often failed to render or rendered in an unrecognizable fashion. For example, as seen in \cref{fig:chip-issue}, the bottoms of chip models have significant holes. On the other hand, it is easier to generate 3D models of larger items with more texture such as chicken strips or a chicken wing. Yet, irrespective of texture or size, items that fall apart (have high fragility) throughout the entire data collection process also led to poor model renderings. Such an instance is the tuna rice ball. Though the 3D model for one tuna rice ball is successfully created, most of the tuna rice balls failed to capture as the tuna would slip or change shape when the sushi is flipped which resulted in a poor 3D model rendering as seen in \cref{fig:tuna-issue}. Thus, extra care had to be taken during data collection for fragile food items to ensure that a high-quality model could be captured. A generalized summary of properties that contribute to the success of a 3D model rendering along with examples is displayed in Table ~\ref{generalized-model-properties}.

\begin{table}[t]
    \caption{Quantifiers and examples for various properties that contribute to a good quality (green) and poor quality (red) model rendering.}
    \setlength{\tabcolsep}{4pt}
    \centering
    \scalebox{1}{
        \begin{tabular}{|c|c|c|}
            \hline 
             Property & Quantifier & Example
            \\ 
            \hline
            \multirow{2}{6em}{Texture} & \cellcolor{red!25} Low & \cellcolor{red!25} Cheese Block \\ & \cellcolor{green!25} High & \cellcolor{green!25} Granola Bar \\
            \hline
            \multirow{2}{6em}{Volume} & 
            \cellcolor{red!25} Low & \cellcolor{red!25} Grape \\ & \cellcolor{green!25} High & \cellcolor{green!25} Apple \\ 
            \hline
            \multirow{2}{6em}{Thickness} & \cellcolor{red!25} Low & \cellcolor{red!25} Potato Chip \\ & \cellcolor{green!25} High & \cellcolor{green!25} Salad Chicken Strip \\ 
            \hline
            \multirow{2}{6em}{Fragility} & \cellcolor{green!25} Low & \cellcolor{green!25} Chicken Wing \\ & \cellcolor{red!25} High & \cellcolor{red!25} Tuna Rice Ball \\
        \hline
        \end{tabular}}
    \label{generalized-model-properties}
\end{table}

\section{NutritionVerse-3D Dataset}
\begin{figure}[h]
    \begin{center}
        \includegraphics[width=.8\linewidth]{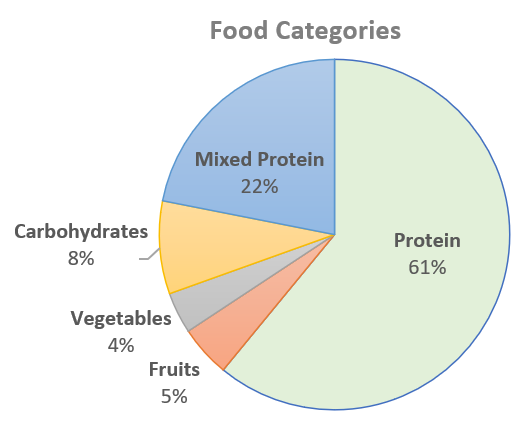}
    \end{center}
    \caption{Count of 3D food models in each food category in the NutritionVerse-3D dataset.}
    \label{food-category-count}
\end{figure}

105 food models comprising of 20 unique types are created successfully using the pipeline proposed in Section 2 and are saved in the OBJ and PLY file formats, two of the most widely used file formats for 3D models \cite{overview-3d-data-formats}. Provided along with the models are their associated weight, food name, and nutritional value. The total number of food models per category is shown in ~\cref{food-category-count} with mixed protein referring to food items (e.g., tuna rice ball) that contain almost equal amounts of protein and other categories such as carbohydrates. 

Along with each model are the nutritional information for the food item. A sample of the metadata file is shown in Table ~\ref{sample-metadata-file}.

\begin{table*}[th]
\caption{Sample portion of the metadata file (iron, magnesium, potassium, sodium, vitamin D, and vitamin B12 values are also available).}
\begin{tabular}{|l|r|r|r|r|r|r|}
\hline
\textbf{item\_id} &
  \textbf{food\_weight\_grams} &
  \textbf{calories} &
  \textbf{fat} &
  \textbf{carbohydrates} &
  \textbf{protein} &
  \textbf{calcium} \\
  \hline
id-10-nature-valley-granola-bar-18g & 18 & 86.40 & 3.87 & 11.59 & 1.62 & 0.01\\
id-11-red-apple-145g                & 145 & 85.55 & 0.29 & 20.39 & 0.39 & 0.01  \\
id-12-carrot-9g                    & 9 & 3.69  & 0.02 & 0.86  & 0.08 & 0.00 \\
id-13-salad-beef-strip-1g          & 1 & 2.15  & 0.08 & 0.00  & 0.32 & 0.00 \\
id-14-salad-beef-strip-7g          & 7 & 15.05 & 0.58 & 0.00  & 2.26 & 0.00 \\
\hline
\end{tabular}
 \label{sample-metadata-file}
\end{table*}

\begin{figure}[h]
    \begin{center}
        \includegraphics[width=\linewidth]{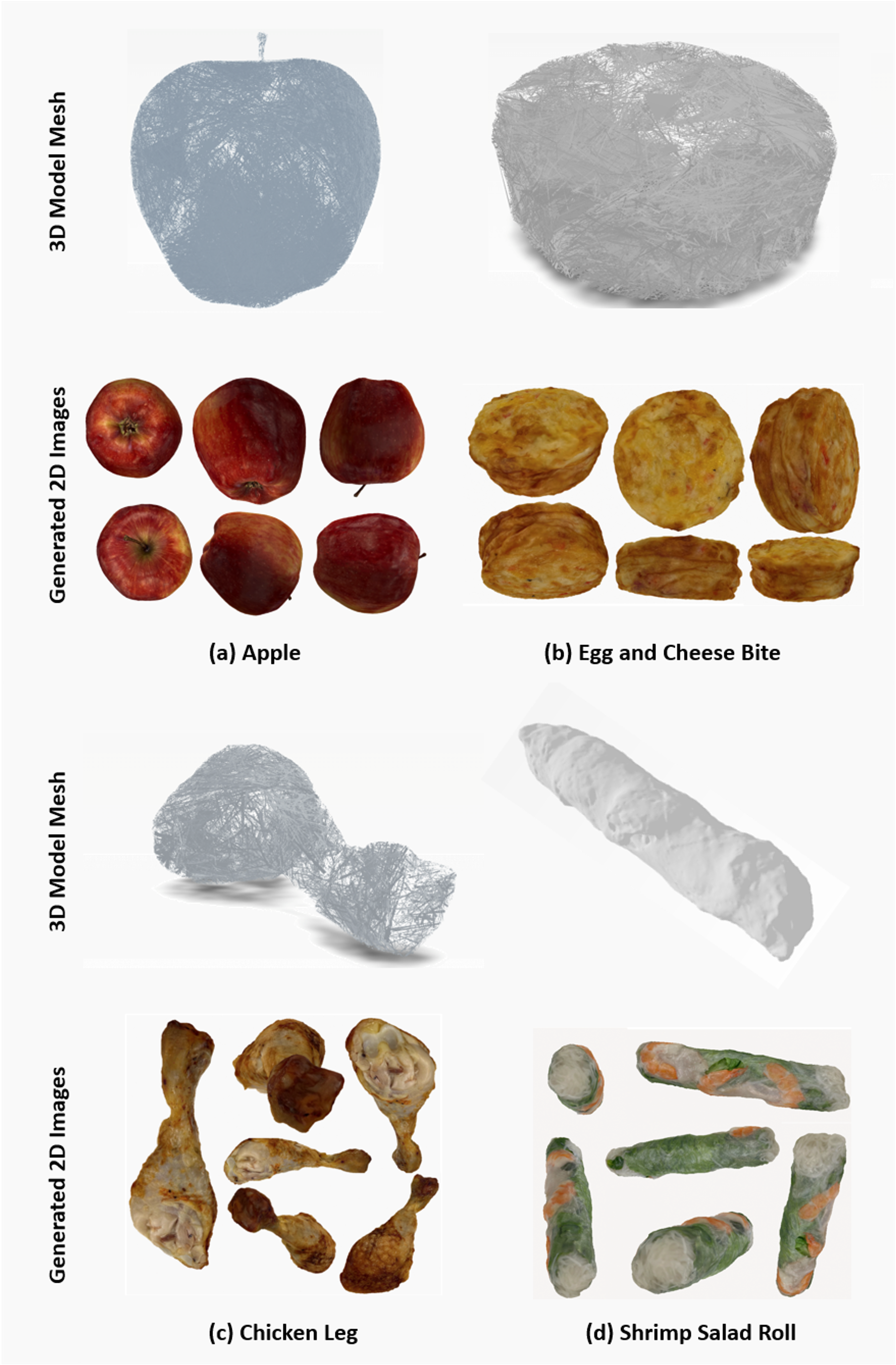}
    \end{center}
    \caption{Example of 2D images obtained from 3D models of an apple, an egg and cheese bite, a chicken leg, and a shrimp salad roll.}
    \label{3d-model-mesh-view-synthesis}
\end{figure}

Another benefit of these 3D food models is that they allow for view synthesis. Examples of leveraging view synthesis with a 3D food model are shown in \cref{3d-model-mesh-view-synthesis} for an apple, an egg and cheese bite, a chicken leg, and a shrimp salad roll. View synthesis is utilized in these figures as the postprocessed sample of generated 2D images includes angles of the food that were not captured in the initial data collection process. As a result, similar 2D images obtained by postprocessing 3D food models extend beyond the fixed camera angles used in the data collection process to reduce imbalance or bias towards a certain viewing angle.

\begin{figure*}[h]
    \begin{center}
        \includegraphics[width=\linewidth]{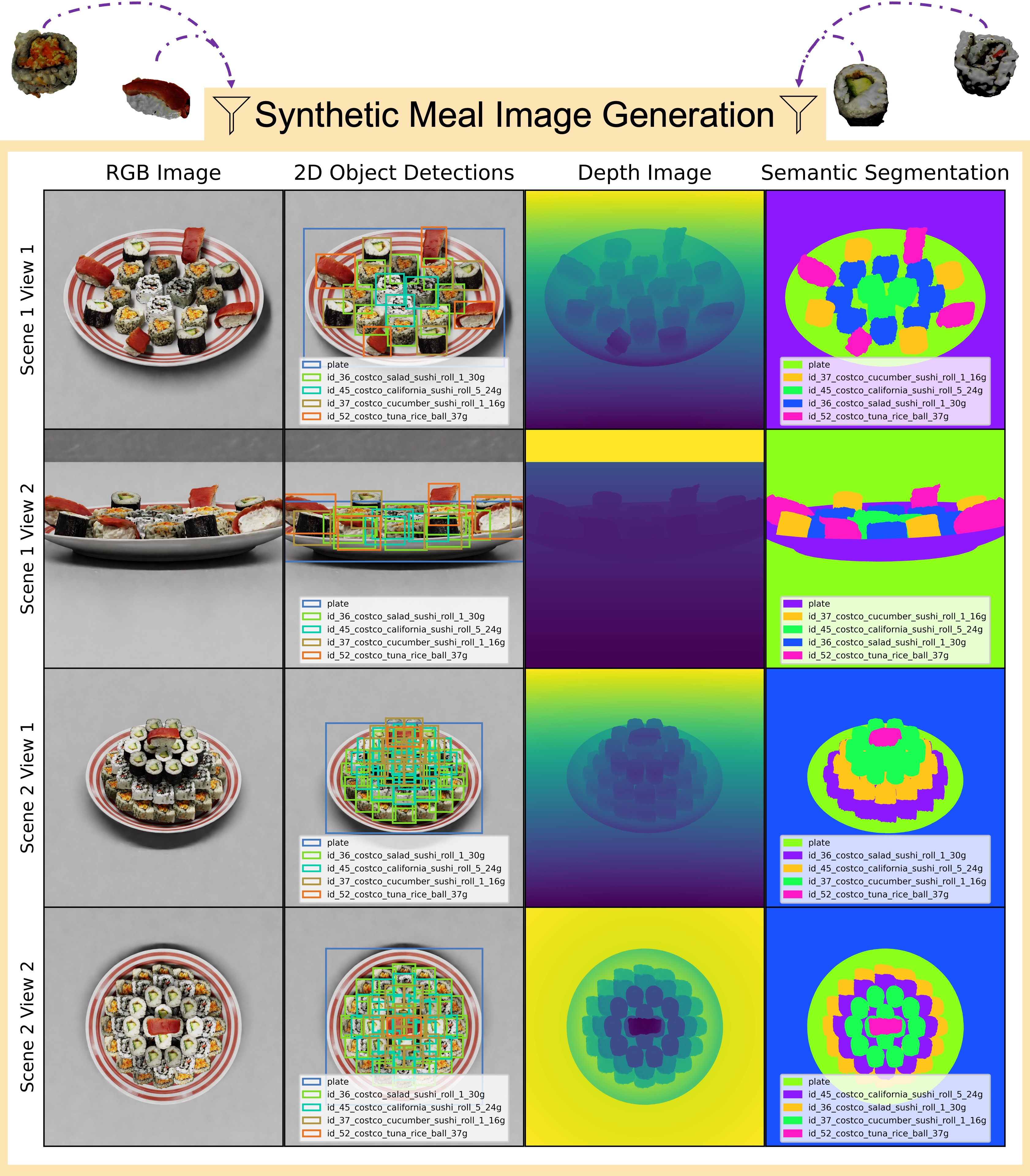}
    \end{center}
    \caption{Automatic synthetic meal generation of various 3D food scenes with multi-modal image data (e.g., RGB and depth data) and annotation metadata (e.g., object detection annotations and segmentation annotations) using the proposed NutritionVerse-3D dataset. Nvidia Omniverse is used to easily capture camera perspectives and lighting conditions that are both diverse and realistic, including scenes where large parts of the dish may be out of focus (e.g. Scene 1, View 2) or occluded by other items (e.g. Scene 2, View 2).}
    \label{fig:synthetic_meal}
\end{figure*}

\section{Synthetic meal image generation}

Given the proposed NutritionVerse-3D dataset with photorealistic 3D food models, one can now automatically generate synthetic multi-modal meal datasets consisting of different sensing modalities (e.g., RGB and depth data) and different forms of annotation (e.g., object detection annotations and segmentation annotations) along with nutritional metadata for the given generated meal.  An illustration of this automatic synthetic meal generation process is shown in \cref{fig:synthetic_meal}.  First, a random number of food items with desired food categories are selected from the NutritionVerse-3D dataset.  The 3D models of the selected food items are then procedurally generated on a virtual plate within Nvidia Omniverse, a platform for digital twin simulation.  Omniverse is then used to generate an RGB image and the corresponding depth image of the scene from a random camera angle, as well as the associated object detection bounding boxes and semantic segmentation annotations.  Finally, the nutritional metadata from the dataset is then used to map each food item in the synthetic scene as well as compute total nutrition information for the meal.

\section{Conclusion}
In this paper, we introduced NutritionVerse-3D, a large-scale high-quality, high-resolution dataset of 105 3D food models in conjunction with their associated weight, food name, and nutritional value. The methodology to collect this dataset was also presented along with the encountered challenges to develop the pipeline. Leveraging the 3D models in the dataset, 3D food scenes can be generated and when coupled with automated view synthesis algorithms, an infinite number of 2D images can be obtained from any angle. Such an approach would allow for a more representative and unbiased image dataset that can be used to develop an effective model for nutritional intake tracking for older adults.

\section{Future Work}
Further studies can be conducted using NutritionVerse-3D to generate an assortment of 3D food scenes and an automated collection of a variety of 2D images from different angles, quality, and lighting condition. A major challenge with creating a food dataset is accounting for numerous dish combinations and layouts. Having 3D models of individual food items permits efficient swapping and the automated assembly of a variety of dish combinations. As an example, by having the individual pieces of a salad, one could assemble various types of salad without actually having to obtain and image each salad type. Furthermore, substitution of sides in a dish would be as easy as swapping the 3D food models. Such substitution could be easily automated using food categories and adding constraints to ensure realistic food renderings. 

\section*{Acknowledgments}
The authors thank National Research Council Canada and the Aging in Place (AiP) Challenge Program. The authors also thank their partners in the Kinesiology and Health Sciences department Dr. Heather Keller, Dr. Sharon Kirkpatrick, and Meagan Jackson. The authors also thank undergraduate research assistants Tanisha Nigam and Komal Vachhani.

{\small
\bibliographystyle{ieeetr}
\bibliography{main}
}

\end{document}